\documentclass{amia}
 \pdfoutput=1

\usepackage{booktabs}
\usepackage{multirow}
\usepackage{wrapfig}
\usepackage{graphicx}
\usepackage{hyphenat}
\usepackage{amsfonts}
\usepackage{amsmath}
\usepackage{xcolor}

\usepackage[superscript,nomove]{cite}
\newcommand{\redtext}[1]{#1}
\graphicspath{{./figure/}}

\begin{document}

\title{Generalizing through Forgetting - Domain Generalization for Symptom Event Extraction in Clinical Notes}

\author{Sitong Zhou, MS$^1$, Kevin Lybarger, PhD$^2$, Meliha Yetisgen, PhD$^1$, Mari Ostendorf, PhD$^1$}

\institutes{$^1$University of Washington, Seattle, WA, USA, $^2$George Mason University, Fairfax, VA, USA}

\maketitle

\noindent{\bf Abstract}

\textit{
Symptom information is primarily documented in free-text clinical notes and is not directly accessible for downstream applications. To address this challenge, information extraction approaches that can handle clinical language variation across different institutions and specialties are needed. 
In this paper, we present domain generalization for symptom extraction using pretraining and fine-tuning data that differs from the target domain in terms of institution and/or 
specialty and patient population. 
We extract symptom events using a transformer-based joint entity and relation extraction method. To reduce reliance on domain-specific features, we propose a domain generalization method that dynamically masks frequent symptoms words in the source domain. Additionally, we pretrain the transformer language model (LM) on task-related unlabeled texts for better representation. Our experiments indicate that masking and adaptive pretraining methods can significantly improve performance when the source domain is more distant from the target domain.}


\section*{Introduction}
\label{section:introduction}

    
The assessment of symptoms, which are physical or mental problems that patients experience, is critical for disease diagnosis \cite{Cahan2017ALH} and epidemic forecasting \cite{Rossman2020AFF}. Symptoms are often documented in free-text clinical notes in Electronic Health Records (EHRs) and are not directly accessible as structured data for downstream applications such as disease prediction or healthcare outcome management. To address this need, a variety of entity and relation extraction methods \cite{Luan2019AGF,Eberts2020SpanbasedJE,Zhong2020AFE,Lybarger2021ExtractingCD,Li2020ChineseCN} have been published to help automate the extraction of symptoms from clinical notes. 

Clinical language varies across different institutions and specialties. Additionally, symptom phrase distribution shifts and symptom contexts vary across different patient populations. Considering the costs of manual annotations from medical experts, creating gold standard annotations to capture the entire clinical note distribution within EHRs does not scale. One common approach is to sample notes to annotate only from the domains of immediate interest (e.g., COVID-19 \cite{Lybarger2021ExtractingCD}) and apply the trained model to other target domains. In such cross-domain settings, we observed a significant extraction performance drop when a symptom extraction model trained on clinical notes of COVID patients is applied to clinical notes of cancer patients. \cite{grace}

    

        

Current state-of-the-art information extraction methods \cite{Eberts2020SpanbasedJE,Zhong2020AFE,Ye2022PackedLM} use transformer-based language models (LM) \cite{Devlin2019BERTPO,Liu2019RoBERTaAR} that are pretrained on large scale corpora. However, the pretrained corpus may not cover the target task data distribution; for example, Bio+Clinical BERT \cite{Alsentzer2019PubliclyAC} is trained based on clinical notes from patients admitted to intensive care units (ICUs) from one medical center. Our patient population is not limited to ICU patients and the notes are from two other medical centers.
These types of data distribution differences are often referred to as ``domain mismatch,'' but the term ``domain'' is not always well defined.  For purposes of this work, a domain is defined as a collection of clinical notes associated with a particular set of patient cohorts, types of notes, and medical centers.


    
In order to avoid annotating \redtext{ training data in every new target domain}, we train the model on the source domain where annotations exist and evaluate the trained model in the target domain. To improve the cross-domain performance, we improve the text representation by pretraining LM on relevant texts to our task based on the success of adaptive pretaining \cite{Gururangan2020DontSP,Han2019UnsupervisedDA}. Furthermore, we challenge the model by masking symptom phrases that are frequently seen in the source domain to reduce the influence of source domain features and force it to learn more generalizable contextual patterns.


In this work, we focus on cross-domain symptom event extraction from outpatient clinical notes from University of Washington Medical Center (UWMC) to cancer treatment ones from Seattle Cancer Care Alliance (SCCA). Our contributions are summarized below. 
(1) We propose a domain generalization method that randomly masks frequent symptoms in the source domain during fine-tuning to encourage the model to give more attention to the context.
(2) We observe that source-target domain differences impact the effects of both adaptive pretraining and symptom masking methods, and the source domain more distant to the target receives more benefits. 
(3) We find that masking frequent symptoms in the source domain helps detect symptom triggers \footnote{Triggers are symptom phrases themselves that indicate the events, and described by other arguments. More details in the Task section. } that are more likely labeled as non-triggers in the source domain.

\section*{Related work}
\label{section:related_work}

%
    

    We can approach our task as event extraction, where each symptom event is identified by a trigger and characterized by multiple arguments linked to the trigger. Events can be extracted by joint entity and relation extraction methods. 
    DYGIE++ \cite{Wadden2019EntityRA} uses graphs to propagate contextual information on top of the transformer encoder. SpERT \cite{Eberts2020SpanbasedJE} employs more lightweight entity and relationship classifiers without compromising performance. PURE \cite{Zhong2020AFE} uses different transformer encoders for entity and relation classification and fuses entity information into relation extraction by inserting markers. Besides, some works formulate joint entity and relation extraction as table filling \cite{Wang2020TwoAB} and question answering\cite{Li2019EntityRelationEA}. We choose SpERT because it is among the state-of-arts on SciERC \cite{Luan2018MultiTaskIO}, and has been successful in the clinical application \cite{Lybarger2022ExtractingRF}. We use SpERT for all experimentation, in order to focus the investigation on domain generalization, not extraction architecture design.

    Pretraining a transformer LM on large corpora is necessary for it to succeed in downstream tasks \cite{ Brown2020LanguageMA,Devlin2019BERTPO}. It is beneficial to train on pretraining corpora that are closer to the task domain \cite{Wadden2019EntityRA, Eberts2020SpanbasedJE}. A variety of off-the-shelf domain-specific LMs are available, including SciBERT \cite{Beltagy2019SciBERTAP} for scientific articles, BioBERT \cite{Lee2020BioBERTAP} and PubMedBERT \cite{Gu2022DomainSpecificLM} for biomedical literature, and Bio+Clinical BERT \cite{Alsentzer2019PubliclyAC} for clinical notes. The Bio+Clinical BERT is pre-trained on MIMIC III \cite{Johnson2016MIMICIIIAF}, which consists of clinical notes of patients admitted to intensive care units (ICU) from a single medical center. The off-the-shelf clinical LM may not be adequate for representing our task since our data cover a broader range of patient populations from different medical systems. Adaptive pretraining on task-relevant data has shown benefits for downstream tasks \cite{Gururangan2020DontSP, Han2019UnsupervisedDA}. Thus, we further pretrain the Bio+Clinical BERT using unlabeled texts relevant to our task.
    

    
    This paper focuses on the case where the source domain for training mismatches the target domain for evaluation.
    A model can overly remember features specific to the source domain and not perform well on the target domain. This problem can be tackled by domain adaptation (DA) \cite{Ramponi2020NeuralUDSurvey} and domain generalization (DG) \cite{Wang2021GeneralizingTU, Shen2021TowardsOG}. The difference between DA and DG is that DA uses target domain data. The key to success is to learn domain-invariant features that are informative to task classifiers. Some methods learn domain-invariant features using domain-adversarial training \cite{Ganin2016DomainAdversarialTO, Naik2020TowardsOD,Du2020AdversarialAD}; some reweight or select source domain training data similar to target domain data \cite{Ruder2017DataSS, Aharoni2020UnsupervisedDC,  Naik2020TowardsOD, Iter2022TheTO,Iter2021OnTC}; some manipulate datasets by pseudo-labeling \cite{Ruder2018StrongBF, Zou2018UnsupervisedDA} or generating new examples \cite{Calderon2022DoCoGenDC, Volpi2018GeneralizingTU}; some create auxiliary tasks \cite{Amin2021T2NERTB} such as optimizing LM objectives for target domains \cite{Peng2015NamedER, Vu2020EffectiveUD, Karouzos2021UDALMUD}, solving jigsaw puzzles \cite{Carlucci2019DomainGB}, and predicting coarse entity types \cite{Liu2020CoachAC}.

To the best of our knowledge, the data manipulation DG method we propose is new to entity and relation extraction tasks. This method is compatible with arbitrary models. Instead of augmenting data and introducing noise in most data manipulation methods \cite{Ruder2018StrongBF,Calderon2022DoCoGenDC}, we mask trigger phrases randomly from the labeled data to challenge the model.

\section*{Task} \label{section:task}

\begin{figure}[ht!]

    \centering
    \includegraphics[scale=0.26]{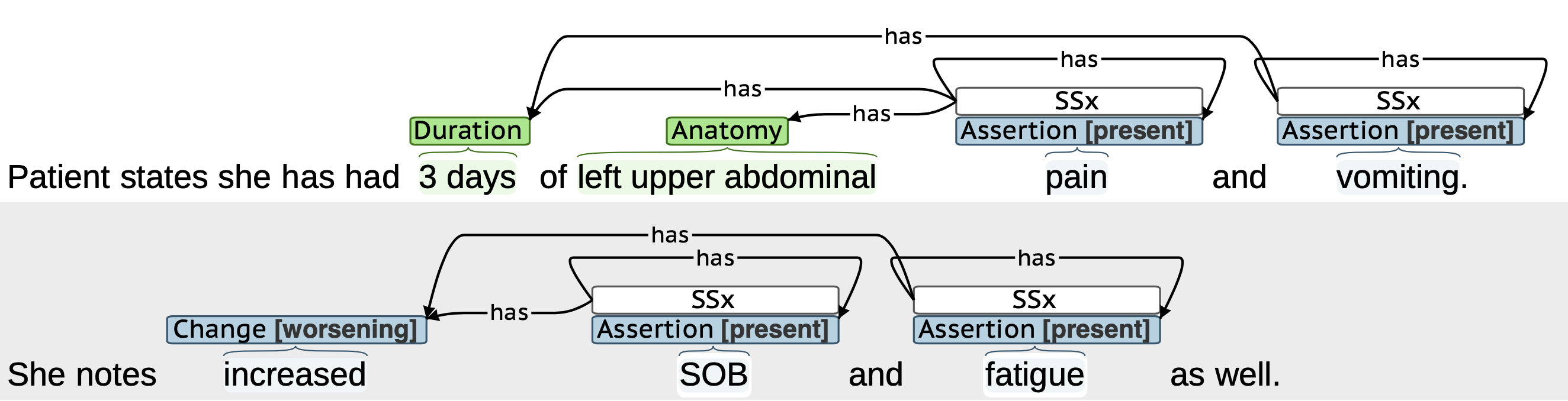}
    \caption{Annotated samples show entity spans with relations linking arguments to symptom triggers. Span colors denote symptoms (white), labeled arguments (blue), and span-only arguments (green). \textit{Assertion} spans match triggers.}
    \label{figure:event_schema}
\end{figure}

In this work, we explore a symptom extraction task, where extracted symptoms are represented using an event-based annotation schema that is tailored to clinical text. We adopt the event schema from COVID-19 Annotated Clinical Text (CACT) Corpus \cite{Lybarger2021ExtractingCD}, where \textbf{triggers} are symptom entities (e.g., pain, vomiting); argument entities are divided into \textbf{labeled arguments} (e.g., \textit{Assertion}) and \textbf{span-only arguments} (e.g., \textit{Anatomy}) depending on whether they contain entity subtype labels. Details of the symptom event schema are presented in Table \ref{table:event_schema} and example annotations are presented in  \ref{figure:event_schema}.

\begin{table}
\centering

\scalebox{0.7}{
\begin{tabular}{cccc}
\toprule
\textbf{Event type, }                                 & \textbf{Argument type}                 & \textbf{Argument subtypes}                                                                    & \textbf{Span examples}              \\ \toprule
\multirow{7}{*}{Symptom}                 & SSx - Trigger\textsuperscript{*}          & --                                                                                    & ``cough," ``shortness of breath"                          \\ \cmidrule{2-4} 
                                            & Assertion\textsuperscript{*}      & \{\nohyphens{present, absent, possible, conditional, hypothetical, not patient}\}         & ``admits," ``denies"                                      \\ \cmidrule{2-4} 
                                                    & Change                            & \{no change, worsened, improved, resolved\}                                               & ``improved," ``continues"                                 \\ \cmidrule{2-4}                                                     
                                                    & Severity                          & \{mild, moderate, severe\}                                                                & ``mild," ``required ventilation"                          \\ \cmidrule{2-4} 
                                                    & Anatomy                           & --                                                                                    & ``chest wall," ``lower back"                              \\ \cmidrule{2-4} 
                                                    & Characteristics                   & --                                                                                    & ``wet productive," ``diffuse"                             \\ \cmidrule{2-4} 
                                                    & Duration                          & --                                                                                    & ``for two days," ``1 week"                        \\ \cmidrule{2-4}                                                                                                                                                                                     
                                                    & Frequency                         & --                                                                                    & ``occasional," ``chronic"                                 \\ \bottomrule 
\end{tabular}}

\caption{Entity types and subtypes for symptom events. Entities can be grouped into triggers, labeled arguments, and span-only arguments. Labeled arguments, including \textit{Assertion}, \textit{Change} and \textit{Severity}, have subtypes \cite{Lybarger2021ExtractingCD}. }
\label{table:event_schema}
\end{table}


Our task is similar to joint entity and relation extraction tasks \cite{Luan2018MultiTaskIO}, but we require trigger entities to be present and arguments to be linked to triggers. We focus on span-level prediction; let $\mathbb{X}$ and $\mathbb{S}$ denote the space of all input sequences and their enumerated spans, and let $Y_e$ and $Y_r$ represent all entity types, and all relation types, including the negative types when labels are missing. For each input sentence $x$ and its labels $y_e$ and $y_r$, the entity classifier predicts the entity type of span $s$ with $f_e:\mathbb S, \mathbb X \to\mathbb Y_e$, where $ f_e(s, x) = y_e(s)$, and the relation classifier predicts whether span $s$ and span $s'$ are linked using $f_r:\mathbb (S,S) \to\mathbb Y_r  $ where $ f_r(s, s', x) = y_r(s,s')$.


After prediction, events are constructed for each trigger entity, adding non-trigger entities as event arguments if they are linked to the trigger head.

The \textit{Assertion} argument includes subtype labels that indicate whether the identified symptom is \textit{present}, \textit{absent}, \textit{conditional}, etc. For \textit{absent} symptoms, there are consistent negation cues, like ``denies'' or ``no.'' While there are affirming cues, like ``reports'' or ``has'' for \textit{present} symptoms, the \textit{present} subtype is often implied by a lack of negation cues. To provide the \textit{Assertion} span classifier with a more consistent span representation, we replaced each  \textit{Assertion} span with the trigger span in each event. The extraction model treats each trigger and its \textit{Assertion} argument as the same entity which has the trigger span and the \textit{Assertion} subtype. We unmerged triggers and \textit{Assertion} arguments into entities with the same spans in post-processing for evaluation purpose; \textit{Assertion} entities keep the subtype labels, and trigger entities do not. 



\section*{Datasets} 
In this paper, we use the following four resources. 

The COVID-19 Annotated Text Corpus ({\em COVID}) contains clinical notes of 230,000 outpatient clinic patients treated at UWMC between May and June 2020\cite{Lybarger2021ExtractingCD}. {\em COVID} contains telephone encounter notes, progress notes, and emergency department notes from UWMC. 

The Lung Cancer Annotated Text Corpus ({\em Lung}) \cite{grace} was constructed from clinical notes of 4,673 lung cancer patients diagnosed between 2012 and 2020. {\em Lung} includes outpatient progress notes, admission notes, emergency department notes, and discharge notes that were created 24 months prior to cancer diagnosis at UWMC. 

The Ovarian Cancer Annotated Text Corpus ({\em Ovarian}) \cite{grace} is based on the clinical notes of 173 ovarian cancer patients diagnosed between 2012-2021.  {\em Ovarian}  includes outpatient progress notes, admission notes, emergency department notes, discharge notes, and gynecology notes created 12 months prior to cancer diagnosis at UWMC.

The Post-Cancer Diagnosis Annotated Text ({\em Post}) was built using clinical notes from 2,968 prostate cancer patients and 1,222 Diffuse Large B cell lymphoma patients diagnosed between 2007-2021. {\em Post} contains notes from SCCA after cancer diagnosis, including notes from urology, oncology, hematology, surgery, radiation oncology, and palliative care.  

All four resources include a variety of clinical notes from UWMC and SCCA. Lung, Ovarian and Post datasets are all related to cancer patients. Lung and Ovarian share the same group of annotators. COVID has the most labeled examples. For each domain, we retrieved unlabeled clinical notes from the same distribution as the labeled ones. Table \ref{table:stats_data} provides more details about the data size, with inter annotator agreement for the three most frequent entity types.
\begin{table}[!ht]
    \centering
\begin{tabular}{lrrrcl}
\toprule
     Source &   \begin{tabular}[c]{c} Labeled  \\ Reports   \end{tabular} &  \begin{tabular}[c]{c} Labeled  \\ Sentences   \end{tabular} &  \begin{tabular}[c]{c} Trigger  \\ Instances   \end{tabular} & \begin{tabular}[c]{c} Inter-Annotator  \\     Trigger / Assertion /Anatomy     \end{tabular}   & \begin{tabular}[c]{c} Unabeled  \\ Reports   \end{tabular}  \\
\midrule
      COVID &    1028 &     89573 &    16966 &   86/83/81 &        87723 \\
       Lung &     145 &     22190 &     3937 &   74-83/70-79/71-79   &        281493 \\
    Ovarian &     100 &     15031 &     2477 &    79-82/71-81/64-79     &     29298 \\
Post (Test) &     200 &     28764 &    5619  & 80-83/73-74/74-76 &               -- \\
\bottomrule
\end{tabular}

    \caption{Dataset information for labeled training data in the source domains, labeled test data in the target domain, and unlabeled data in both the source and target domains. }
    \label{table:stats_data}
\end{table}



\begin{figure}[h!]
    \centering
    \includegraphics[scale=0.55]{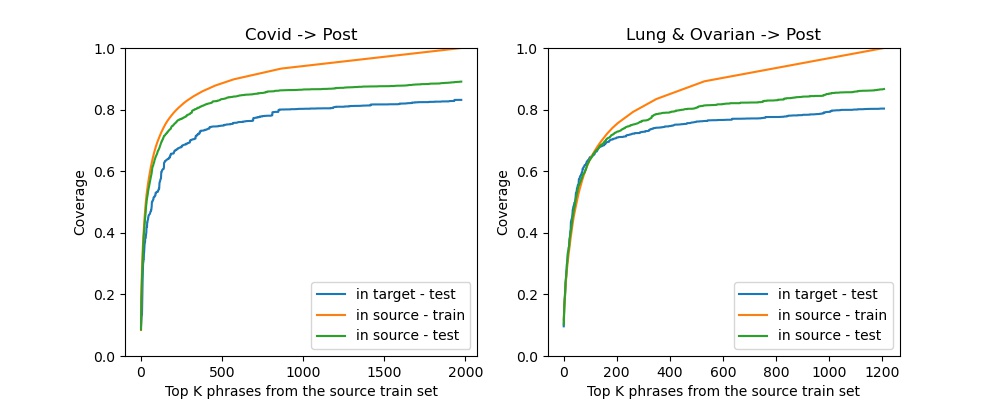}
    \caption{Domain discrepancy: source-target domain differences in terms of trigger coverage.}
    \label{figure:domain_discrepancy_coverage}
\end{figure}

In this study, we choose Post as the target domain, and we use two source sets: COVID and the combination of Lung \& Ovarian (due to their small size). 
All source domains differ from the target in institution as well as diagnosis stage and diseases types of patients. The Lung \& Ovarian source is similar to the target domain as it also involves cancer patients, but it does not relate to prostate or lymphoma cancers. 
The impact of these domain differences between source and target domains is illustrated in Figure \ref{figure:domain_discrepancy_coverage}, which plots the coverage of triggers across domains. 


\section*{Methods}
\label{section:method}



In our cross-domain setting, we assume labeled training data is available for the source domain only, and that the source differs from the target domain. The baseline cross-domain model leverages a standard configuration of a pretrained transformer incorporated into a model that is trained on the labeled source domain data. 
The experiments look at two strategies for improving cross-domain performance: (i) the use of unlabeled data in the source domain for adaptive pretraining, and (ii) the use of dynamic masking in the supervised training stage.
We explore the benefits of the two approaches alone and in combination for different degrees of domain mismatch. Both adaptive pretraining and supervised training are on source domain data; inference at test time is on target domain data.




\subsection*{Baseline}

\begin{figure}[ht!]
\centering
    \includegraphics[scale=0.037]{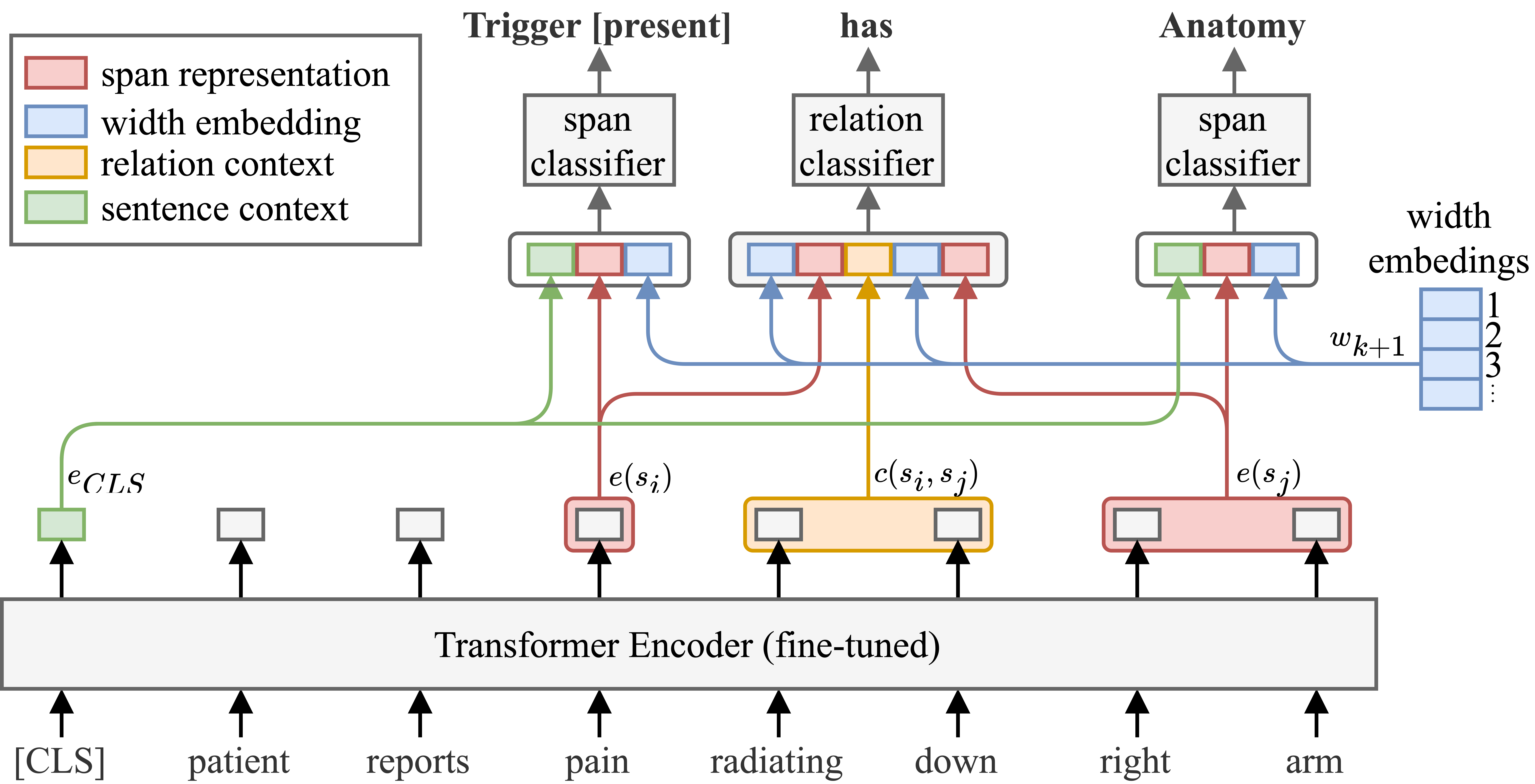}
    \caption{SpERT Model Achitecture based on work  \cite{Eberts2020SpanbasedJE,Lybarger2022ExtractingRF} }
    \label{figure:spert_method}
\end{figure}

All experiments are run on the transformer-based SpERT \cite{Eberts2020SpanbasedJE}, which is one of state-of-art models on the SciERC benchmark \cite{Luan2018MultiTaskIO}. According to the original SpERT work \cite{Eberts2020SpanbasedJE} and a clinical application based on SpERT \cite{Lybarger2022ExtractingRF}, the model architecture is shown in Figure \ref{figure:spert_method}. SpERT contains light-weighted entity and relation classifiers on top of the BERT encoder \cite{Devlin2019BERTPO}. Specifically, we use a clinical version of BERT, Bio+Clinical BERT \cite{Alsentzer2019PubliclyAC}. The \textbf{entity classifier} predicts enumerated spans at the subtype level. To represent each span, SpERT groups the encoder outputs of tokens within the span by max pooling, and concatenates it with the sentence context and span size embeddings. To predict whether a relation exists between any pair of entity spans, we represent the span pair using their entity representations plus the max pooled encoder vectors of the words between the two spans. The \textbf{relation classifier} classifies the relation representations into a binary outcomes.



  
To train the model, we minimize the cross entropy (CE) loss between predictions and labels, sampling non-entity spans and span-pairs without relations as negative examples. To normalize the outputs of the multiclass entity classifier, we use the softmax function, while for the binary relation classifier, we use sigmoid.
For input sentence $x$,   labels for entities $y_e$ and relations $y_r$, let $f_{e}$ and \redtext{$f_{r}$} denotes the model outputs of entities and relations, the objective is 
$L_{Joint} = L_{Entity} + L_{Relation}$, where
$L_{Entity} = \sum_{s\in Spans(x)}  CE({f}_{e}(s, x),y_{e}(s))$,$ \quad L_{Relation} = \sum_{(s,s')\in Span Pairs(x)} CE({f}_{r}(s, s', x),y_{r}(s, s'))$


\subsection*{Adaptive Pretraining}
The baseline model is initialized with Bio+Clinical BERT \cite{Alsentzer2019PubliclyAC} pretrained on biomedical research articles and clinical notes. However, the pretraining corpus could fail to represent our task distribution, because its clinical note source, MIMIC-III \cite{Johnson2016MIMICIIIAF} is mostly from ICU patients from one medical center, but we are targeting cancer patients who are not necessarily admitted to ICUs and from another medical center.
To adjust the LM to the target domain, we continue pretraining on unlabeled texts relevant to our task as in the DAPT method \cite{Gururangan2020DontSP}. Pretraining minimizes the masked language model (MLM) objective. 
For any input $x$ with tokens ${t_0, t_1, ... t_{l}}$, we randomly replace  tokens $t_i$ to [MASK] at a fixed rate (15\%). Let the set of masked token indexes in sentence $x$ be $LMMask(x)$ and the sentence after masking be $\tilde{x}$; let the LM $f_{LM}(i, \tilde{x})$ predicts for the original $i$-th token that is masked in the new context $\tilde{x}$, the MLM pretraining objective is $L_{MLM} =   \sum_{i \in LMMask(x)}  CE(f_{LM}(i, \tilde{x}),t_i)$


In adaptive pretraining, we use the combined set of unlabeled texts from COVID, Lung, and Ovarian, with domain indicator tokens at the start of a sequence. We also conducted experiments using unlabeled target domain texts as an additional pretraining corpus, but results are not reported since it was not helpful. \redtext{We adopt the same masking rate of 15\% for the MLM pre-training objective as in the original BERT model\cite{Devlin2019BERTPO}.}

\subsection*{Dynamic Masking}
As shown earlier, domains may differ in the frequency of specific trigger words. We hypothesize that, for very frequent triggers, the model may put too much weight on the word sequence alone vs.\ its context, failing to capture more generalizable contextual cues.  
In this scenario, cross-domain performance might be compromised, especially when symptom phrase distribution shifts and symptom contexts change. The model may fail to detect symptoms with a considerable change in context distributions, as well as symptoms less frequent in the source.
 
We challenge the model  
to learn the contextual information for trigger extraction by masking the frequent symptom phrases seen in training. This is a data manipulation method that changes the labeled source data without modifying the model. First, we create the symptom phrase list based on their frequency ranking in the source domain train set, then search those phrases in the source domain training data, and randomly mask the matched phrases with a fixed probability, regardless of whether they are annotated as triggers or not. 
\redtext{We treated the random masking rate as a tunable hyperparameter and determined the optimal rate for our task is 80\%.}
In order to prevent permanent loss of annotations \redtext{due to aggressive masking}, we use dynamic masking \cite{Liu2019RoBERTaAR} to change random masks every epoch. We train the joint entity and relation model on the masked source domain labeled data using the same $L_{joint}$ objective.

\section*{Experiment}
\label{section:experiment}
\subsection*{Implementation}


Each note was split into sentences using spaCy.\footnote{https://spacy.io/models/en\#en\_core\_web\_sm} To encode the text, we use Bio + Clinical BERT's default tokenizer, which contains 28996 uncased tokens. As in Bio + Clinical BERT, the transformer encoder contains 12 layers and 12 heads. The training batch size is 15. For each model, we train it for 10 epochs.



During pretraining, we truncate the text corpus into chunks of 512 length. For each optimization step, we accumulate sub-batches of size 32 eight times in order to have a larger batch size of 256. The MLM random rate is 15\%. When pretraining on all source domains together, the LM is trained for 31 epochs finishing 67.6 K steps. With the target domain texts added, we pretrain on this larger corpus for 30 epochs, which is equivalent to 126.8 K steps.
To create the symptom phrase lists for masking, we used the top 200 frequent symptom phrases from each source. In order to simplify the list, we keep only phrases with one token and exclude punctuation-only \footnote{Trigger labels on punctuation could come from annotation errors.} or single-character triggers. In training data, we search for those listed tokens, and dynamically mask matched tokens at an 80\% random rate.

All SpERT modeling experiments are conducted on a NVIDA GoForce GPU card with 11.6 GB of memory. A NVIDA A100 GPU with 42.5 GB of memory is used to pretrain the language models.


\subsection*{Evaluation}
The evaluation process is similar to our previous symptom event extraction work \cite{Lybarger2021ExtractingCD}. All triggers and arguments are scored by micro F1 scores. The predicted trigger is considered correct if the trigger type and span are both equal to the gold trigger. The subtype-level entity label and the linked trigger must match the gold labels for a predicted labeled argument to be considered correct. When calculating span-only argument metrics, we count the overlap tokens between each predicted span and the gold span, and we require the two spans to have the same entity type and trigger.
When comparing models, we run 5 random seeds and report the results of a two-sided T-test.

\section*{Results \& Discussion}

\begin{table}[ht]
\centering
\resizebox{\textwidth}{!}
{
\begin{tabular}{rr|llll|llll}
\toprule
 & & \multicolumn{4}{c|}{COVID} & \multicolumn{4}{|c}{Lung \& Ovarian} \\
  Entity & NT & Baseline &      \begin{tabular}[c]{l} w/ \\     Mask    \end{tabular} & \begin{tabular}[c]{c} + Adaptive \\     Pretrain    \end{tabular}  &  \begin{tabular}[c]{l} w/ \\     Mask    \end{tabular} &        Baseline &        \begin{tabular}[c]{l} w/ \\     Mask    \end{tabular} & \begin{tabular}[c]{c} + Adaptive \\     Pretrain    \end{tabular}  & \begin{tabular}[c]{l} w/ \\     Mask    \end{tabular} \\
\midrule
SSx             &  5629 &     77.2 &         78.5** &               78.6* &              \textbf{79.9*} &           79.2 &  78.2- - &       \textbf{79.3} &                        79.1 \\
Assertion       &  5629 &     72.2 &          73.1* &              74.0** &               \textbf{74.9} &  \textbf{75.1} &   74.4- &                75.0 &               \textbf{75.1} \\
Change          &   401 &     55.7 &  \textbf{56.9} &       \textbf{56.9} &               \textbf{56.9} &           47.2 &    44.8 &                46.1 &               \textbf{47.3} \\
Severity        &   291 &     30.7 &           30.4 &                29.6 &              \textbf{33.3*} &           40.7 &  36.4- - &       \textbf{42.2} &                        38.5 \\
Anatomy         &  3007 &     55.3 &           56.7 &                57.2 &               \textbf{59.4} &           60.7 &  58.5- - &      \textbf{63.2*} &                        61.4 \\
Characteristics &  1250 &     22.5 &           22.8 &     \textbf{26.8**} &                        25.8 &           15.1 &    16.6 &                16.3 &               \textbf{17.7} \\
Duration        &   837 &     46.5 &           44.1 &                46.3 &               \textbf{48.4} &           29.9 &    32.5 &                30.4 &               \textbf{34.2} \\
Frequency       &   301 &     45.9 &           45.6 &                44.9 &               \textbf{46.6} &           24.7 &    24.9 &      \textbf{27.3*} &                        26.6 \\
\midrule
SSx (Precision)             &   &  \textbf{90.1} &         89.4- - &              89.4- - &                        88.9 &  \textbf{84.1} &  82.5- &                83.1 &                        83.0 \\
SSx (Recall)           &   &     67.5 &  69.9** &               70.1* &              \textbf{72.6*} &           74.8 &    74.3 &      \textbf{75.9*} &                        75.6 \\
\bottomrule
\end{tabular}
}
 \caption{Entity micro F1 scores and trigger (SSx) precision and recall scores for both sources.  **(- -) and *(-) indicate significant gain(loss) relative to contrasting condition at p-value smaller than 0.01 and 0.05, respectively. We have four different versions of models: Baseline, Baseline with dynamic masking, Adaptive Pretrainining, and Adaptive Pretraining with dynamic masking. "NT" stands for "number of true labels," which refers to the count of gold labels.
}
    \label{table:results_all_arguments_all}
\end{table}

Table \ref{table:results_all_arguments_all} summarizes the cross-domain results on the trigger and argument extraction. The Lung \& Ovarian source domain performs better than the COVID domain for the three most frequent entity types. 

\textbf{Adaptive pretraining benefits performance for both source domains.} For the COVID source model, the gains are significant for trigger (\textit{SSx}), \textit{Assertion} and \textit{Characteristics}. For the models trained on the Lung \& Ovarian source, \redtext{The gains are significant in} \textit{Anatomy} and \textit{Frequency} arguments.


\redtext{\textbf{Dynamic masking benefits performance for the models trained on the COVID domain, but has varying effects for different source domains.} Specifically,  dynamic masking improves the trigger extraction of COVID models with statistical significance, for both scenarios with and without adaptive pretraining, but hurts the Lung \& Ovarian source model performance when no adaptive pretraining is applied. The improvement for COVID models on trigger F1 scores is primarily driven by the increase in recall. There is a trade-off between recall increase and precision decrease, but recall dominates the change. However, for Lung \& Ovarian source models, without adaptive pretraining, the precision of trigger extraction significantly decreases after dynamic masking. The varying effects can likely be attributed to differences in the distributions of symptom phrases and symptom trigger contexts across domains. 
}

The combination of masking and pretraining makes it possible to take advantage of the greater amount of labeled data, giving performance that is similar to or better than the best result from the Lung \& Ovarian source model for all entity types except \textit{Severity} and \textit{Anatomy}. 
The best COVID source model gives a result close to the human annotator agreement for triggers (80-83) and \textit{Assertion} (73-74) entities. 


\begin{table}[!ht]
    \centering
    \scalebox{1.0}{

 \begin{tabular}{lllllll}
\toprule
               &                 &      top 20 &     top 20-40 &    top 40-60 &    top 60-80 &   top 80-100 \\
Source & Version &                &                  &                 &                 &                \\
\midrule
COVID & Baseline &           95.4 &             89.9 &            83.9 &            83.1 &           82.6 \\
               &      w/ masking &           95.7 &  \textbf{93.1**} &            84.2 &          87.1** &           82.7 \\
               & + Adaptive Pretraining &           96.3 &             90.2 &            84.2 &            82.6 &  \textbf{83.8} \\
               &      w/ masking &  \textbf{96.5} &           92.8** &  \textbf{85.8*} &  \textbf{87.2*} &           83.0 \\
Lung \& Ovarian & Baseline&  \textbf{97.4} &             77.8 &            81.9 &   \textbf{76.6} &           73.8 \\
               &      w/ masking &         96.2- - &             77.5 &   \textbf{82.5} &          72.9- - &           73.6 \\
               & + Adaptive Pretraining &  \textbf{97.4} &             77.4 &            80.6 &            76.2 &           74.8 \\
               &      w/ masking &         96.6- -  &    \textbf{78.3} &            82.3 &           73.6 - &  \textbf{76.0} \\
\bottomrule
\end{tabular}
}
    \caption{Trigger F1 results on subsets of symptom phrases grouped by source domain frequency. **(- -) and *(-) indicate significant gain(loss) relative to contrasting condition at p-value smaller than 0.01 and 0.05, respectively.}
    \label{table:results_infrequent_bin}
\end{table}

In order to better understand the effect of trigger frequency in the source domain, we looked at performance for different sets of triggers grouped by source frequency. 
Table~\ref{table:results_infrequent_bin} provides F1 results for the different training configurations.
For COVID, masking has little impact on the performance for the most frequent source triggers, but significantly benefits the triggers in the top 20-40 and top 60-80 groups. 
In contrast, masking significantly hurts the most frequent source triggers when the source is Lung \& Ovarian.

\begin{figure}[h!]
    \centering
    \includegraphics[scale=0.56]{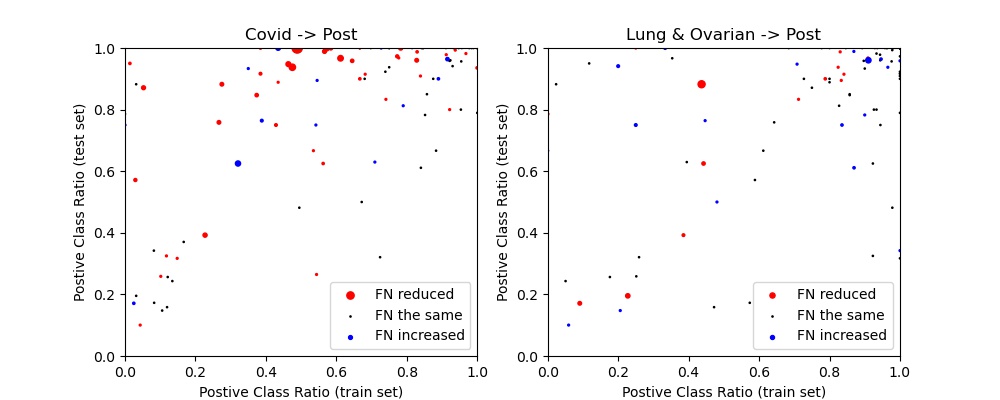}
    \caption{Relative frequency of a positive trigger label in the test vs.\ training set for the 100 most frequent triggers in the test set. Size of the points indicates the absolute value of the symptom phrase false negative (FN) change after masking. Red indicates that false negatives are reduced; blue indicates no change or an increase.}
    \label{figure:analysis_ner_r}
\end{figure}

 \textbf{Masking benefits phrases that are are more likely associated with negative contexts in the source than in the target}. A trigger phrase that is frequently annotated as a non-trigger in the source can lead to false negatives in the target domain. 
To study how the negative context shift affects masking, we plot positive class ratio of trigger phrases on both source and target domains, which is the likelihood of the phrases annotated as triggers (Figure \ref{figure:analysis_ner_r}). We size and color each point according to the absolute value and sign of the false negative change. We show only false negatives because the trigger detection improvement after masking is due to the reduced false negatives. Most phrases with reduced false negatives are above the diagonal, in other words, the phrases improved after masking are more likely to be annotated as non-triggers in the source domain than in the target. The COVID source has more trigger phrases overly associated with negative contexts and sitting above the diagonal, which can be compensated for by masking.


\section*{Conclusion}
\label{section:conclusion}
We use adaptive pretraining and our proposed dynamic masking method to improve symptom event extraction under cross-domain settings without using additional labeled data.
Both methods significantly benefit when the source domain is more distant from the target, and achieve trigger F1 scores close to the human annotator agreement.
The dynamic masking method improves the detection of symptoms that are less likely to be 
 annotated as symptom triggers in the source domain. For future work, understanding the type of domain discrepancy could guide us in selecting domain generalization (or adaptation) methods.
 
 While we primarily focus on analyzing the symptom phrase characteristics when describing the shifts, other factors such as the amount of source data, and annotation consistency can also affect domain generalization. Future work should explore ways to analyze and handle different types of shifts, and we recommend cross-domain studies controlling other factors such as annotation quality.


\section*{Acknowledgements} This work was supported by NIH/NCI (1R01CA248422-01A1 and 1R21CA258242-01).


\bibliography{main}

\begin{thebibliography}{10}

\bibitem{Cahan2017ALH}
Cahan A, Cimino JJ.
\newblock A Learning Health Care System Using Computer-Aided Diagnosis.
\newblock Journal of Medical Internet Research. 2017;19.

\bibitem{Rossman2020AFF}
Rossman H, Keshet A, Shilo S, Gavrieli A, Bauman T, Cohen O, et~al.
\newblock A framework for identifying regional outbreak and spread of COVID-19
  from one-minute population-wide surveys.
\newblock Nature Medicine. 2020:1  4.

\bibitem{Luan2019AGF}
Luan Y, Wadden D, He L, Shah A, Ostendorf M, Hajishirzi H.
\newblock A general framework for information extraction using dynamic span
  graphs.
\newblock In: NAACL; 2019. .

\bibitem{Eberts2020SpanbasedJE}
Eberts M, Ulges A.
\newblock Span-based Joint Entity and Relation Extraction with Transformer
  Pre-training.
\newblock In: ECAI; 2020. .

\bibitem{Zhong2020AFE}
Zhong Z, Chen D.
\newblock A Frustratingly Easy Approach for Joint Entity and Relation
  Extraction.
\newblock In: NAACL; 2021. .

\bibitem{Lybarger2021ExtractingCD}
Lybarger K, Ostendorf M, Thompson M, Yetisgen M.
\newblock Extracting COVID-19 Diagnoses and Symptoms From Clinical Text: A New
  Annotated Corpus and Neural Event Extraction Framework.
\newblock In: J Biomed Inform; 2021. .

\bibitem{Li2020ChineseCN}
Li X, Zhang H, Zhou XH.
\newblock Chinese clinical named entity recognition with variant neural
  structures based on BERT methods.
\newblock In: J Biomed Inform; 2020. p. 103422.

\bibitem{grace}
Turner G, Lybarger K, Brennan A, Chang E, Dorvall N, Gill J, et~al.
\newblock Domain Adaptation of a Deep Learning Symptom Extractor for Different
  Patient Populations and Clinical Settings.
\newblock In: AMIA Annual Symposium Proceedings; 2021. .

\bibitem{Ye2022PackedLM}
Ye D, Lin Y, Li P, Sun M.
\newblock Packed Levitated Marker for Entity and Relation Extraction.
\newblock In: Assoc Comput Linguist; 2022. .

\bibitem{Devlin2019BERTPO}
Devlin J, Chang MW, Lee K, Toutanova K.
\newblock BERT: Pre-training of Deep Bidirectional Transformers for Language
  Understanding.
\newblock In: NAACL; 2019. .

\bibitem{Liu2019RoBERTaAR}
Liu Y, Ott M, Goyal N, Du J, Joshi M, Chen D, et~al.
\newblock RoBERTa: A Robustly Optimized BERT Pretraining Approach.
\newblock ArXiv. 2019;abs/1907.11692.

\bibitem{Alsentzer2019PubliclyAC}
Alsentzer E, Murphy JR, Boag W, Weng WH, Jin D, Naumann T, et~al.
\newblock Publicly Available Clinical BERT Embeddings.
\newblock In: NAACL; 2019. .

\bibitem{Gururangan2020DontSP}
Gururangan S, Marasovi{\'c} A, Swayamdipta S, Lo K, Beltagy I, Downey D, et~al.
\newblock Don’t Stop Pretraining: Adapt Language Models to Domains and Tasks.
\newblock In: Assoc Comput Linguist; 2020. .

\bibitem{Han2019UnsupervisedDA}
Han X, Eisenstein J.
\newblock Unsupervised Domain Adaptation of Contextualized Embeddings: A Case
  Study in Early Modern English.
\newblock In: EMNLP-IJCNLP; 2019. .

\bibitem{Wadden2019EntityRA}
Wadden D, Wennberg U, Luan Y, Hajishirzi H.
\newblock Entity, Relation, and Event Extraction with Contextualized Span
  Representations.
\newblock In: EMNLP-IJCNLP; 2019. .

\bibitem{Wang2020TwoAB}
Wang J, Lu W.
\newblock Two Are Better than One: Joint Entity and Relation Extraction with
  Table-Sequence Encoders.
\newblock In: EMNLP; 2020. .

\bibitem{Li2019EntityRelationEA}
Li X, Yin F, Sun Z, Li X, Yuan A, Chai D, et~al.
\newblock Entity-Relation Extraction as Multi-Turn Question Answering.
\newblock In: Assoc Comput Linguist; 2019. .

\bibitem{Luan2018MultiTaskIO}
Luan Y, He L, Ostendorf M, Hajishirzi H.
\newblock Multi-Task Identification of Entities, Relations, and Coreference for
  Scientific Knowledge Graph Construction.
\newblock In: EMNLP; 2018. .

\bibitem{Lybarger2022ExtractingRF}
Lybarger K, Damani A, Gunn ML, Uzuner {\"O}, Yetisgen-Yildiz M.
\newblock Extracting Radiological Findings With Normalized Anatomical
  Information Using a Span-Based BERT Relation Extraction Model.
\newblock In: AMIA Annual Symposium proceedings.. vol. 2022; 2022. p. 339-48.

\bibitem{Brown2020LanguageMA}
Brown TB, Mann B, Ryder N, Subbiah M, Kaplan J, Dhariwal P, et~al.
\newblock Language Models are Few-Shot Learners.
\newblock In: NeurIPS; 2020. .

\bibitem{Beltagy2019SciBERTAP}
Beltagy I, Lo K, Cohan A.
\newblock SciBERT: A Pretrained Language Model for Scientific Text.
\newblock In: EMNLP; 2019. .

\bibitem{Lee2020BioBERTAP}
Lee J, Yoon W, Kim S, Kim D, Kim S, So CH, et~al.
\newblock BioBERT: a pre-trained biomedical language representation model for
  biomedical text mining.
\newblock Bioinformatics. 2020;36:1234  1240.

\bibitem{Gu2022DomainSpecificLM}
Gu Y, Tinn R, Cheng H, Lucas MR, Usuyama N, Liu X, et~al.
\newblock Domain-Specific Language Model Pretraining for Biomedical Natural
  Language Processing.
\newblock ACM Transactions on Computing for Healthcare (HEALTH). 2022;3:1  23.

\bibitem{Johnson2016MIMICIIIAF}
Johnson AEW, Pollard TJ, Shen L, wei H~Lehman L, Feng M, Ghassemi MM, et~al.
\newblock MIMIC-III, a freely accessible critical care database.
\newblock Scientific Data. 2016;3.

\bibitem{Ramponi2020NeuralUDSurvey}
Ramponi A, Plank B.
\newblock Neural Unsupervised Domain Adaptation in NLP—A Survey.
\newblock In: Proceedings of the 28th International Conference on Computational
  Linguistics; 2020. .

\bibitem{Wang2021GeneralizingTU}
Wang J, Lan C, Liu C, Ouyang Y, Qin T.
\newblock Generalizing to Unseen Domains: A Survey on Domain Generalization.
\newblock In: IJCAI; 2021. .

\bibitem{Shen2021TowardsOG}
Shen Z, Liu J, He Y, Zhang X, Xu R, Yu H, et~al.
\newblock Towards Out-Of-Distribution Generalization: A Survey.
\newblock ArXiv. 2021;abs/2108.13624.

\bibitem{Ganin2016DomainAdversarialTO}
Ganin Y, Ustinova E, Ajakan H, Germain P, Larochelle H, Laviolette F, et~al.
\newblock Domain-Adversarial Training of Neural Networks.
\newblock In: J. Mach. Learn. Res.; 2016. .

\bibitem{Naik2020TowardsOD}
Naik A, Ros'e CP.
\newblock Towards Open Domain Event Trigger Identification using Adversarial
  Domain Adaptation.
\newblock In: Assoc Comput Linguist; 2020. .

\bibitem{Du2020AdversarialAD}
Du C, Sun H, Wang J, Qi Q, Liao J.
\newblock Adversarial and Domain-Aware BERT for Cross-Domain Sentiment
  Analysis.
\newblock In: Assoc Comput Linguist; 2020. .

\bibitem{Ruder2017DataSS}
Ruder S, Ghaffari P, Breslin JG.
\newblock Data Selection Strategies for Multi-Domain Sentiment Analysis.
\newblock ArXiv. 2017;abs/1702.02426.

\bibitem{Aharoni2020UnsupervisedDC}
Aharoni R, Goldberg Y.
\newblock Unsupervised Domain Clusters in Pretrained Language Models.
\newblock In: Assoc Comput Linguist; 2020. .

\bibitem{Iter2022TheTO}
Iter D, Grangier D.
\newblock The Trade-offs of Domain Adaptation for Neural Language Models.
\newblock In: Assoc Comput Linguist; 2022. .

\bibitem{Iter2021OnTC}
Iter D, Grangier D.
\newblock On the Complementarity of Data Selection and Fine Tuning for Domain
  Adaptation.
\newblock ArXiv. 2021;abs/2109.07591.

\bibitem{Ruder2018StrongBF}
Ruder S, Plank B.
\newblock Strong Baselines for Neural Semi-Supervised Learning under Domain
  Shift.
\newblock In: Assoc Comput Linguist; 2018. .

\bibitem{Zou2018UnsupervisedDA}
Zou Y, Yu Z, Kumar BVKV, Wang J.
\newblock Unsupervised Domain Adaptation for Semantic Segmentation via
  Class-Balanced Self-training.
\newblock In: ECCV; 2018. .

\bibitem{Calderon2022DoCoGenDC}
Calderon N, Ben-David E, Feder A, Reichart R.
\newblock DoCoGen: Domain Counterfactual Generation for Low Resource Domain
  Adaptation.
\newblock In: Assoc Comput Linguist; 2022. .

\bibitem{Volpi2018GeneralizingTU}
Volpi R, Namkoong H, Sener O, Duchi JC, Murino V, Savarese S.
\newblock Generalizing to Unseen Domains via Adversarial Data Augmentation.
\newblock In: NeurIPS; 2018. .

\bibitem{Amin2021T2NERTB}
Amin S, Neumann G.
\newblock T2NER: Transformers based Transfer Learning Framework for Named
  Entity Recognition.
\newblock In: EACL; 2021. .

\bibitem{Peng2015NamedER}
Peng N, Dredze M.
\newblock Named Entity Recognition for Chinese Social Media with Jointly
  Trained Embeddings.
\newblock In: EMNLP; 2015. .

\bibitem{Vu2020EffectiveUD}
Vu TT, Phung DQ, Haffari G.
\newblock Effective Unsupervised Domain Adaptation with Adversarially Trained
  Language Models.
\newblock In: EMNLP; 2020. .

\bibitem{Karouzos2021UDALMUD}
Karouzos CF, Paraskevopoulos G, Potamianos A.
\newblock UDALM: Unsupervised Domain Adaptation through Language Modeling.
\newblock In: NAACL; 2021. .

\bibitem{Carlucci2019DomainGB}
Carlucci FM, D'Innocente A, Bucci S, Caputo B, Tommasi T.
\newblock Domain Generalization by Solving Jigsaw Puzzles.
\newblock IEEE/CVF Conference on Computer Vision and Pattern Recognition
  (CVPR). 2019:2224-33.

\bibitem{Liu2020CoachAC}
Liu Z, Winata GI, Xu P, Fung P.
\newblock Coach: A Coarse-to-Fine Approach for Cross-domain Slot Filling.
\newblock In: Assoc Comput Linguist; 2020. .

\end{thebibliography}
\end{document}